# Dynamic replanning in uncertain environments for a sewer inspection robot

**Oliver Adria[1]; Hermann Streich[2]; Joachim Hertzberg[3]**
[1,2,3]Fraunhofer Institute for Autonomous Intelligent Systems,
Sankt Augustin, Germany
e-mail : [first name].[last name]@ais.fraunhofer.de

*Abstract :* *The sewer inspection robot MAKRO is an autonomous multi-segment robot with worm-like shape driven by wheels. It is currently under development in the project MAKRO-PLUS. The robot has to navigate autonomously within sewer systems. Its first tasks will be to take water probes, analyze it onboard, and measure positions of manholes and pipes to detect polluted-loaded sewage and to improve current maps of sewer systems. One of the challenging problems is the controller software, which should enable the robot to navigate in the sewer system and perform the inspection tasks autonomously, not inflicting any self-damage. This paper focuses on the route planning and replanning aspect of the robot. The robot's software has four different levels, of which the planning system is the highest level, and the remaining three are controller levels each with a different degree of abstraction. The planner coordinates the sequence of actions that are to be successively executed by the robot.*

## 1. Introduction

Germany's sewers have a total length of more than 400,000 km and have been continuously extended over the last decades. All sewers are required to be inspected periodically by the municipalities. Currently, inspection is done by a human operator, controlling a cable driven robot which carries a video camera. Many such systems are on the market and currently in use all over the world, e.g. (IBAK), (Aries), (Gejos).

MAKRO (Rome, E. et al., 1999) is an articulated inspection robot of worm-like shape and is designed for autonomous navigation in sewer pipes of 30 to 60 cm width. Its case design, consisting of six segments connected by five motor-driven active joints, allows for simultaneously climbing a step and turning, e.g., at a junction consisting of a 600 millimeter pipe and a branching 300 millimeter pipe with equal top levels (Kepplin, V. et al., 1999). MAKRO's autonomy and its kinematical abilities extend its potential mission range enormously, compared to conventional inspection equipment that is limited by cable and poor kinematics. MAKRO carries all the needed resources on-board. Standard NiCad batteries provide the power for its 21 motors, the sensors, and the electronics, including an industry standard PC/104+ computer system and seven microcontrollers, allowing for an uptime of about two hours. A prototype (MAKRO 1.1, see figure 1) of the robot has been built and is currently used for various experiments. In the new project MAKRO-PLUS a new robot is under development, which shall move in "in use" sewer systems. Two application modules are under development for MAKRO-PLUS: One for exact measuring purposes of manholes and pipes and another module for on-board water analysis.

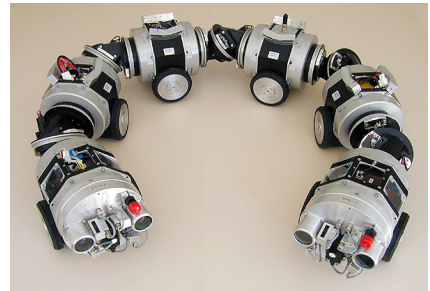

Fig. 1. MAKRO 1.1

This paper focuses on the planning and re-planning system of the robot. Dynamic re-planning is necessary because at any time during execution of its tasks, the robot might unexpectedly run into problems. So to achieve autonomy, the robot has to react to these situations. On encountering problems it either a) tries to complete as many tasks as possible from its tasks list or b) goes to a safety point, where it can be safely recovered by a human operator.

The next chapter will first give an introduction on the robot, its mechanical and electronical features. What follows is a short overview of the robot controller software approach. Then we give details on the workings of the planning system used in the robot. At the end we present a sample mission, then the status of our current work.

The project MAKRO-PLUS has been partially supported by the German Federal Ministry of Research and Technology (BMBF, No. 02WK0259), project partners being rhenag, FZI, FZK, and Inspector System Rainer Hitzel.

## 2. The Robot

### 2.1. Application Area

The robot should be able to navigate autonomously in sewer pipe systems, and able to climb up and down steps up to 30 cm high. It has to find and identify manholes and junctions, must be able to turn within manholes up to 90 degrees. The surface of pipes may be slippery and obstacles like ingrown roots or parts from damaged pipes may block the robot's path. Since recovering the robot from the middle of a long pipe is very expensive and impractical, the robot should always be able to navigate itself to a manhole from which it can be recovered, even in critical situations. Remote operation by radio link within a sewer pipe is not possible due to high signal damping and severe constraints on feasible antenna geometries. If the robot can no longer move, it should send out signals in order to be localized.

A sewer system is a highly structured environment. It consists of very few types of objects, namely pipes and manholes. Pipes mainly differ in length and diameter (and material, which fortunately is not important for the software - after we did some tests on different surfaces we discovered that the robot has no difficulty driving on different types of surfaces). Manholes differ in the number of pipes which lets the water in and out, and the geometry. The most important characteristic is the different heights of pipes, resulting in steps the robot has to climb.

In most industrial countries, sewer system measurements are stored in special databases (sewer information systems). The angles of the different pipe connections and other important data are stored here, so the robot doesn't have to roam in an unknown environment but uses the stored data to complete its tasks.

### 2.2. Mechanical features

Each MAKRO segment is propelled by a motor that drives a two-wheeled axis, and each joint connecting two segments contains three motors that allow for rotations around the three axes in space. The joint motors are strong enough to lift two segments of MAKRO's 50 kilograms heavy, 2 meters long body, which is necessary for climbing steps at some pipe junctions. Since it is impossible to do a 180 degree turn inside a pipe and the robot is required to move backwards if its path is blocked, the robot is built symmetrically with identical head segments on both ends so it can effortlessly change from a forward to a backward direction and vice versa.

### 2.3. Electronics

In total seven microcontrollers and a PC104-computer with a 6.5 GB hard disk, frame grabber and Wireless LAN Ethernet interface are on-board. All processors communicate via CAN-Bus. Each joint has three infrared angle decoders. Each segment has odometers for the wheels. The robot also has a temperature sensor for the PC104 unit. In each head four infrared sensors are used for distance measurement to the sides, up and down. An ultrasonic sensor detects obstacles in the frontal direction. Two inclinometers in each head show the vertical position of the robot. A laser-cross/camera-sensor on each head projects a laser cross into the pipe (Kolesnik, M. & Streich, H., 2002). The processing of the grabbed camera image computes the horizontal angle of the robot's head relative to the axis of the pipe. Two of the robot's segments contain NiCad Batteries, which enables the robot to move for about two hours. To localize the robot if it's unable to move anymore, a signal transmitter is on board, which is activated when deemed necessary.

## 3. Software Approach

### 3.1. Introduction

We decided to organize the mission control software in different degrees of abstraction. The problem was to find a suitable granularity between complex inspection orders (missions) and simple subtasks (e.g. checking a single infrared sensor), alternatives consisting of partitioning the problem into manageable units, finding terms and abstractions for semantically related aspects. The result is a software package which consists of four different levels (see figure 2), which is tailored to the application, namely to autonomously control the sewer inspection robot MAKRO.

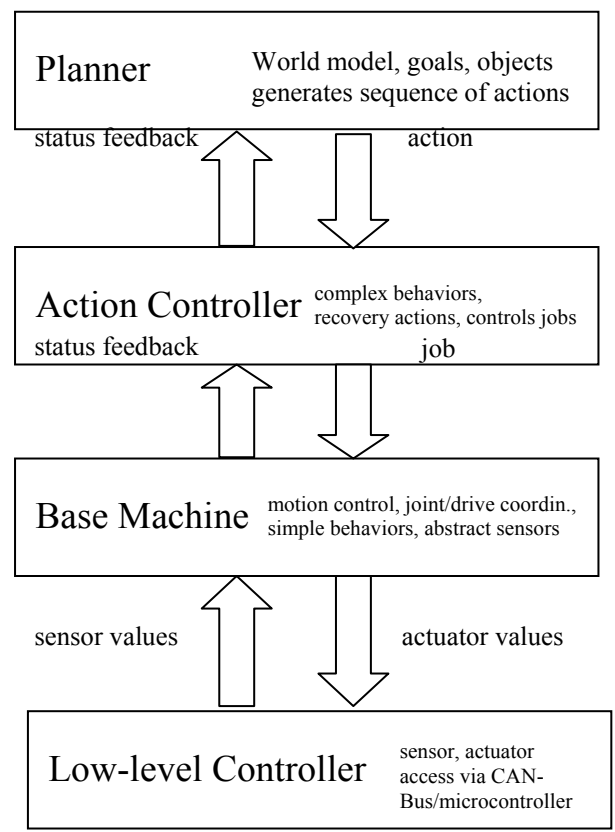

Fig. 2. Our 4-Level Software Approach

*3.2. Overview and Execution Model*

The robot's mission is specified by a human operator who determines where the robot is entered in the sewer system, what inspection tasks the robot has to do at which locations, and where the robot shall be recovered after the inspection tasks have been accomplished. This mission data is transmitted to the robot before the robot is entered into a manhole. As additional information the robot keeps a database, which includes the geometrical data of all pipes and manholes in the inspection area.

It is simple to specify the path the robot has to drive if we have a map. So, we decided to model the world topologically, consisting of manholes and pipes and special locations (inspection points, e.g. for water samples), and then we define some pre-set actions (e.g. moving from Point A to inspection point P). A good AI planning algorithm should be able to generate a sequence of actions so that the robot will reach its specified goals.

The highest level of the 4 software levels used by the robot is the Planner Level. This level generates a sequence of actions that are then successively executed by the next lower level.

The lower 3 levels of the software approach are controller levels with different degrees of abstraction.

Execution of the first tasks begins when the Planner level finds a solution and generates a symbolic plan. This plan consists of a sequence of actions that are recognized by the next lower level: The Action Controller. An action is a function that might look like "DriveToManhole (direction, distance, type, size, speed …)". In case something goes wrong, e.g. there is an obstacle blocking the path of the robot, the Action Controller will respond to this by sending out the appropriate response/error code to the Planner, and the Planner can respond accordingly, depending on the type of error response the Action Controller has sent. The planner might re-route the robot to another path or just send it to the next available manhole for immediate recovery. The Action Controller discriminates three types of errors: Blockage (the robot's path is blocked), danger (robot is in a position where it can damage itself and thus might become immobile) and malfunction (software or hardware error). In case of malfunction we might lose partial control of the robot, so the recovery scheme is for the robot to reboot itself and hope that the fault was temporary, so the robot can continue its mission.

An action itself consists of smaller operations, specified as "jobs". Jobs are implemented one level below the Action Controller level: "Base Machine".

The Base Machine sequentially executes its jobs, just like the Action Controller executes its actions. If the Base Machine should have trouble executing any of its jobs, an error command will be sent to the Action Controller, and the Action Controller will react accordingly. A job is a simple operation which might look like: "DriveForward (speed, distance)".

The 4th and lowest level of the software is the Low-Level Controller. This is the interface to all sensors and actuators in the robot. These are accessible via CAN-bus and are directly connected to the microcontrollers. The low-level controller is provided by our project partner FZI. It runs distributedly on the seven microcontrollers of the robot.

**4. The (Re-)Planning System**

*4.1. Overview*

In the following we will use the word "Planner" or "Planning System" when referring to the planning system implemented in the robot (see figure 3). The AI Planner that is integrated in the Planning System is referred to as "AI planner" or "planning algorithm".

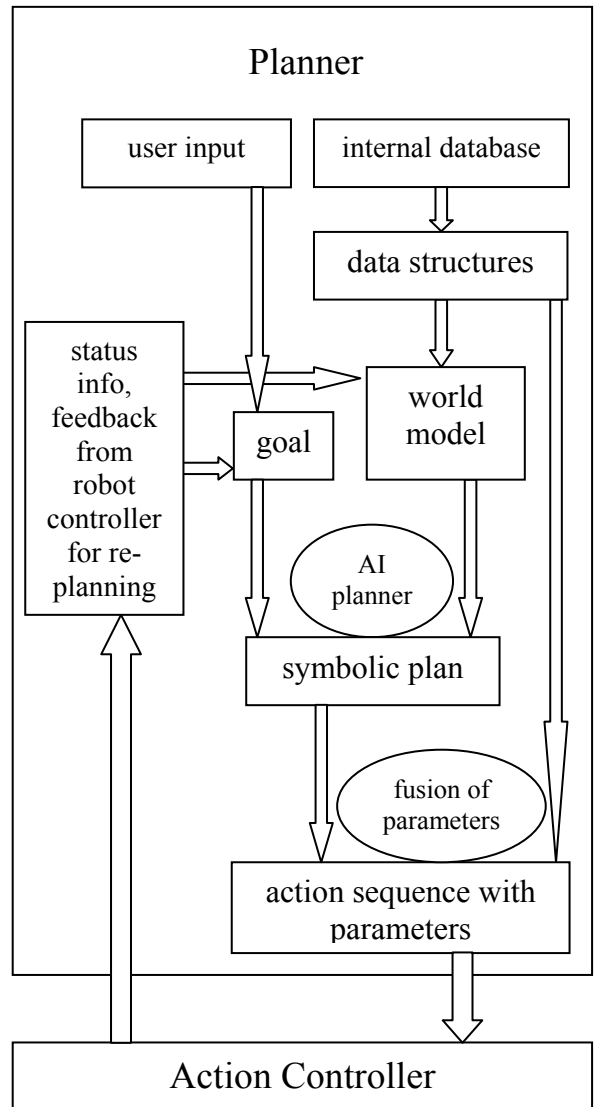

Fig. 3. Planning System Software Architecture

The Planning System of the robot needs two inputs. These are 1) user input for the goals and the specific tasks that are to be completed and 2) the internal database of the sewer where the robot will be working in. The internal database is then parsed into data structures and through the data structure we create a world model. The world model and the set tasks/goals are then fed into the AI planner. The output is a symbolic plan of the robot's tasks. This symbolic plan has no quantitative values so it has to be fused with the data structures to get the parameters needed. This 'fused plan' is a sequence of actions; these are fed into the Action Controller which executes them successively. If an error occurs, the Action Controller will inform the Planning System, and the Planning System will react accordingly by changing the world model and the robot's goals to create a new plan that has been adapted to the modified world model/task list.

*4.2. Internal database and data structures*
The data we need is stored in a sewer database system (*Kanalinformationssystem,* KIS). These databases contain data on the pipes and manholes e.g. unique id number, sizes, connections, coordinates, last date of inspection, type of inspection, just to name a few. Much of this data is not really needed by the robot, so we need to parse this data into the internal database format that the robot understands.
The current internal database used in MAKRO's Planning System is an ASCII text. We set the format ourselves; it includes the manhole's diameter, connections and the gross connection angles between the different pipes. The pipe info stored is the length and the diameter. The ASCII text/database is then subsequently parsed into C++ data structures for easier processing.

*4.3. User Input*
The user has to input the goals and the specific tasks which need to be completed by the robot. There are 3 different tasks that the robot can execute. These are 1) go to a specified point in the sewer system, 2) scan/inspect area and 3) get water sample. This is all that is needed.
In order for the user to input, the user will see a 2-D representation of the sewer system on a PDA (personal digital assistant) and can then click on the spots where the tasks need to be done in the sewer. When user input and database information is received, the robot is on its own and starts working autonomously.

*4.4. World model and goals*
The world model and goals need to be understood by the AI Planner. There is a standard syntax that was set for the International Planning Competition (IPC) called PDDL (Ghallab, M. et al., 1998). So goals and world model are parsed to the correct syntax of PDDL since this is the syntax understood by the AI planner used in the robot's Planning System.

*4.5. The AI Planner and the output (symbolic plan)*
The AI planner first parses the world model and goals before calculating a route for the robot. The output is a sequence of actions written as ASCII text into a solution file. These actions have no quantitative values and contain only manhole and pipe IDs. Two of these actions might look like:

DRIVE_MANHOLE_TYPE_4_FROM_3_TO_4 M6 P10 P4 P5 P6
TAKE_WATER_SAMPLE P6

where the $M_i$ denote manholes and the $P_i$ pipes.

*4.6. Fusion of parameters*
So in order for the robot to work with parameters, the symbolic plan and the actions with the manhole and pipe IDs have to be 'fused/matched' with the data structures. These actions with parameters are then fed sequentially to the Action Controller. The Action Controller then executes them one by one.

*4.7. Changing the world model and modifying goals*
MAKRO's main task will be to inspect pipes and manholes for damages and repair necessities. So it will be very common that MAKRO will run into trouble while on one of its inspection tasks. E.g. an obstacle might be blocking the path in one of the pipes. In this case, MAKRO has to react accordingly and will handle this problem by replanning its route. This is done by changing the world model and the goals. E.g. if the obstacle is in pipe P5 connecting manholes M12 and M17, then we can just disconnect the connection between M12 and M17 and then let the AI planner find a new route, if one exists.
Consequently, some goals set by the human operator might not be achievable anymore because of this blocked path. So we also need to modify the goals that are to be accomplished.
We then try to tick off the highest possible number of goals from the goals list. If the pre-set exit point cannot be reached anymore, the robot will have to find a new exit point, e.g., the next available manhole.

**5. Third party software**

The AI planning algorithm used in the robot is from the FF planner version 2.3 (Hoffmann, J. & Nebel, B., 2001). FF is a forward chaining heuristic state space planner. We decided to go with the FF planner because it showed good results in the IPC 2002; it is coded in C (small overhead) and is relatively small in size. At the current state, the AI planning algorithm is a standalone program which the robot's planner calls by system call. The results are written to a file and then parsed.
MCA2 (Modular Controller Architecture) (Scholl, K. et al., 2001) is a C++-based framework for use in robots and other hardware. It simplifies the coding and debugging process tremendously and gives better structure to the overall program; it is network-transparent and gives us real-time capability without much extra effort, letting us focus on the actual code content and robot functionality. MCA2 is used throughout the whole robot from the low-level-controls up to the Planning System.
A part of the robot controller is implemented in sE (synERJY) (Budde, R. & Poigné, A., 2000), a synchronous programming language developed at our institute. We used sE because of its support for building hierarchical finite state automata. Communication is done via signals on a virtual "bus" concept. The sE compiler detects race conditions on valued signals, which eliminates difficult to find software faults.

## 6. Experiments and an example

At our institute we have a concrete test environment above ground. We created an internal database based on this sewer system. Each pipe and each manhole has a unique number. The mission is defined (by the human operator) by specifying the starting point, the end point, and inspection tasks (e.g. get a water sample from a specific location). In the example shown in figure 4, the robot first drives along line 1. It gets a water sample and analyzes it. Then the robot drives to the next goal point and makes an inspection and then drives to the exit manhole, where the robot is recovered.

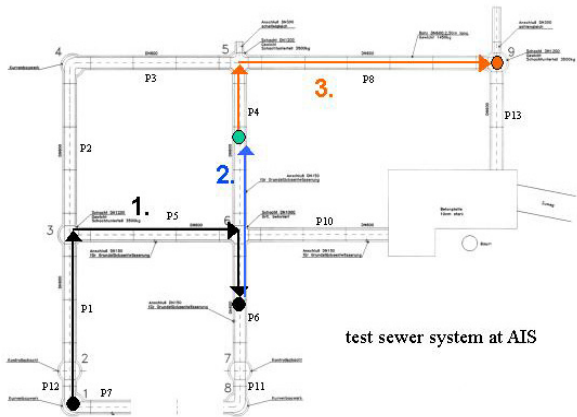

Fig. 4. Test environment at Fraunhofer AIS

The resulting symbolic plan related to figure 4 is as follows:

DRIVE_PIPE_TO_MANHOLE P12 M2
DRIVE_MANHOLE_TYPE_2_FROM_1_TO_2 M2 P12 P1
DRIVE_PIPE_TO_MANHOLE P1 M3
DRIVE_MANHOLE_TYPE_3_TYPE_B_FROM_3_TO_1 M3 P5 P2 P1
DRIVE_PIPE_TO_MANHOLE P5 M6
DRIVE_MANHOLE_TYPE_4_FROM_3_TO_4 M6 P10 P4 P5 P6
TAKE_WATER_SAMPLE P6
DRIVE_PIPE_TO_MANHOLE P6 M6
DRIVE_MANHOLE_TYPE_4_FROM_4_TO_2 M6 P10 P4 P5 P6
DRIVE_PIPE_TO_MANHOLE P4 M5
INSPECT_PIPE P4
DRIVE_MANHOLE_TYPE_4_FROM_4_TO_1 M5 P8 P13 P3 P4
DRIVE_PIPE_TO_MANHOLE P8 M9
DRIVE_MANHOLE_TYPE_3_TYPE_B_FROM_1_TO_2 M9 P8 P9 P14

The execution of this mission has been successfully tested in our simulation (Neuhoefer, J., 2003). In addition to fault-free execution, we also test the behavior if an obstacle is blocking the robot's path. In this situation the responsible recovery module will first drive the robot several centimeters backwards. Then the action of driving forward is retried (phase 1 of the recovery). If the same blockage problem persists, the robot will temporarily lift its head (if possible), hoping the obstacle will somehow disappear (phase 2), e.g. if it was some sort of light waste which did not really obstruct and got flushed away when the robot lifted its head. If that doesn't work, MAKRO tries to push the obstacle out of its way by just driving forward (phase 3). Although this works without any problems in the simulation it is very risky in the real sewer, because it might be possible that the obstruction moves a little and then gets stuck between two segments of the robot. In such delicate cases we will first move the robot safely backward to the next reachable manhole, where it can be recovered.

Within the real test-environment the robot behaves in the same way like in the simulation. Manholes and junctions are detected and verified reliably. Turns up to 90 degrees can be driven without trouble. Figure 5 shows MAKRO 1.1 in the test-environment at our institute, driving trough a manhole in 60 cm diameter pipes.

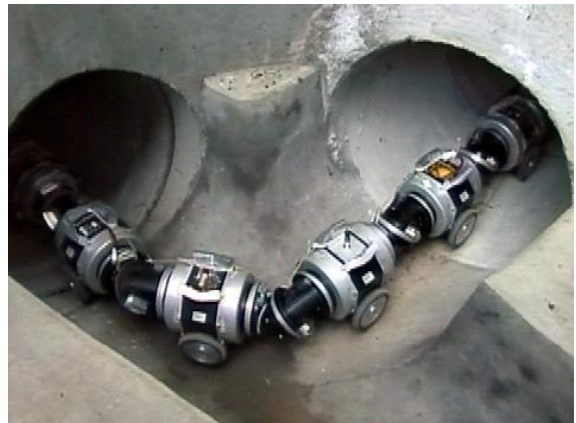

Fig. 5. MAKRO 1.1 turning in AIS test environemt

MAKRO 1.1 is the second prototype we developed. It is water proofed and can be used in dry-weather conditions in real sewage systems. We made tests in Siegburg, a town near the institute, in the sewer system managed by our project partner Rhenag. There we tested the handling of entering and recovering the robot in manholes. In addition we did tests on the remote control possibilities via Wireless LAN Ethernet. It is possible to communicate with the robot if the robot is up to 2 m inside the pipe. This is more than enough for giving execution commands when we are entering and retrieving the robot in manholes and giving the 'go' signal for starting a mission.

## 7. Summary and current state

In this paper we described the software approach in general and the Planning System for the sewer inspection robot MAKRO. Different degrees of abstractions have been defined, which leads to an understandable and extendable structure. MAKRO 1.1 is the prototype for MAKRO PLUS. All the current software tests are based and tested on MAKRO 1.1. The underlying concepts of the robots are identical, but MAKRO PLUS will be water and explosion proof, and will have two special application segments for inspection tasks. The robot will be up to twice as fast (maximum of 60cm/sec), the electronic hardware used (processor, hard disk) is more advanced and powerful. The battery also lasts longer, now able to give power twice as long. At the moment, the mechanical design and the electronic configuration of the robot are at its final stages and a couple of the robot modules are almost finished.

Most parts of the software are running and integrated. A second laser/cam sensor using structured light for obstacle classification and manhole verification is under development.

The Planning System development is in its final stages and nearly finished. Once the interface between the Planning System and the Action Controller is working without software glitches, the next step is to create a user friendly interface. Then the only thing needed is an automated conversion of the data from the huge German sewer database to the internal database needed for the robot, and MAKRO should be fully automated.

## 4. References

Aries Industries, USA, http://www.ariesind.com/, Accessed 2003-12-27

Budde, R.; Poigné, A. (2000) Reactive Control with Simple Synchronous Models, In : Languages, Compilers, and Tools for Embedded Systems, ACM SIGPLAN Workshop LCTES 2000, Vancouver, Canada, Proceedings

Gejos, Germany, http://www.gejos.de/, Accessed 2003-12-27

Ghallab, M.; Howe, A.; Knoblock, C.; McDermott, D.; Ram, A.; Veloso, M.; Weld, D.; and Wilkins, D. (1998). PDDL---The Planning Domain Definition Language. AIPS-98 Planning Committee

Hoffmann, J.; Nebel, B. (2001). The FF Planning System: Fast Plan Generation Through Heuristic Search, in: Journal of Artificial Intelligence Research, Volume 14, Pages 253 - 302.

Kepplin, V.; Scholl, K.-U. & Berns, K. (1999). A mechatronic concept for a sewer inspection robot. In: Proc. IEEE/ASME International Conference on Advanced Intelligent Mechatronics (AIM'99) p. 724–729, IEEE Press, Piscataway, NJ

Kolesnik, M. & Streich, H. (2002). „Visual Orientation and Motion Control of MAKRO - Adaptation to the Sewer Environment". In: B. Hallam, D. Floreano, J. Hallam, and J.-A. Meyer (eds.). From animals to animats 7. Proc. of the 7th Int. Conf. on Simulation of Adaptive Behavior. Cambridge, Mass.: MIT Press, 62-69.

Neuhoefer, J. (2003). Real-Time Simulation of an Inspection Robot with a Commercial Physics Engine, In: Industrial Simulation Conference 2003 / Guerri, Juan Carlos[Hrsg.], p. 375 - 379

Rome, E.; Hertzberg, J.; Kirchner, F.; Licht, U.; Streich, H. & Christaller, Th., (1999). Towards Autonomous Sewer Robots: the MAKRO Project, Urban Water 1, p. 57-70

Scholl, K.-U.; Albiez, J.; Gassmann, B. (2001). MCA – An Expandable Modular Controller Architecture, 3rd Real-Time Linux Workshop, Milano, Italy.

synERJY SE: http://borneo.gmd.de/~ap/sE/index.html, Accessed 2003-12-27